\documentclass[a4paper,conference]{IEEEtran}
\ifCLASSINFOpdf
\else
\fi

\usepackage[utf8x]{inputenc} 
\usepackage{times}
\usepackage{epsfig}
\usepackage{graphicx}
\usepackage{amsmath}
\usepackage{amssymb}
\usepackage{multirow}
\usepackage{multicol}

\usepackage{algorithm}
\usepackage{algorithmicx}
\usepackage{algpseudocode}
\usepackage{amsmath}
\usepackage{float} 
\usepackage{subfigure}

\usepackage{ulem}

\graphicspath{{figs/}}

\begin{document}
%
\title{CCA: Exploring the Possibility of Contextual Camouflage Attack on Object Detection}

\author{\IEEEauthorblockN{Shengnan Hu, Yang Zhang, Sumit Laha, Ankit Sharma and Hassan Foroosh}
\IEEEauthorblockA{Department of Computer Science\\
University of Central Florida\\ 
Orlando, FL, USA\\
\{shengnanhu, yangzhang\}@knights.ucf.edu, laha@cs.ucf.edu,  ankit.sharma285@knights.ucf.edu, hassan.foroosh@ucf.edu}}


%


\maketitle

\begin{abstract}
Deep neural network based object detection has become the cornerstone of many real-world applications. Along with this success comes concerns about its vulnerability to malicious attacks. To gain more insight into this issue, we propose a contextual camouflage attack (\textbf{CCA} for short) algorithm to influence the performance of object detectors. In this paper, we use an evolutionary search strategy and adversarial machine learning in interactions with a photo-realistic simulated environment to find camouflage patterns that are effective over a huge variety of object locations, camera poses, and lighting conditions. The proposed camouflages are validated effective to most of the  state-of-the-art object detectors.
\end{abstract}

\section{Introduction}
Object detection has become a crucial part for many applications, such as autonomous driving \cite{ingle2016tesla}, law enforcement and orbital surveillance \cite{mundhenk2016large}, to name a few. The rapid development of artificial intelligence (AI) and deep neural networks(DNNs) in the past few years boost the advancement of this task. However, state-of-the-art object detectors \cite{he2017mask, liu2016ssd, redmon2018yolov3, ren2015faster} heavily depend on convolutional neural networks (CNNs) \cite{lecun1998gradient} which, unfortunately, have been shown vulnerable to adversarial attack—an adversary which can manipulate the CNNs’ output by adding imperceptible perturbations to an input image \cite{akhtar2018threat}. Moreover, some recent works also show an adversary can even fool object detectors in the real world by adding small distractions to physical objects \cite{athalye2017synthesizing, eykholt2018robust}.

\begin{figure*}[t]
\centering
\includegraphics[width=0.9\textwidth]{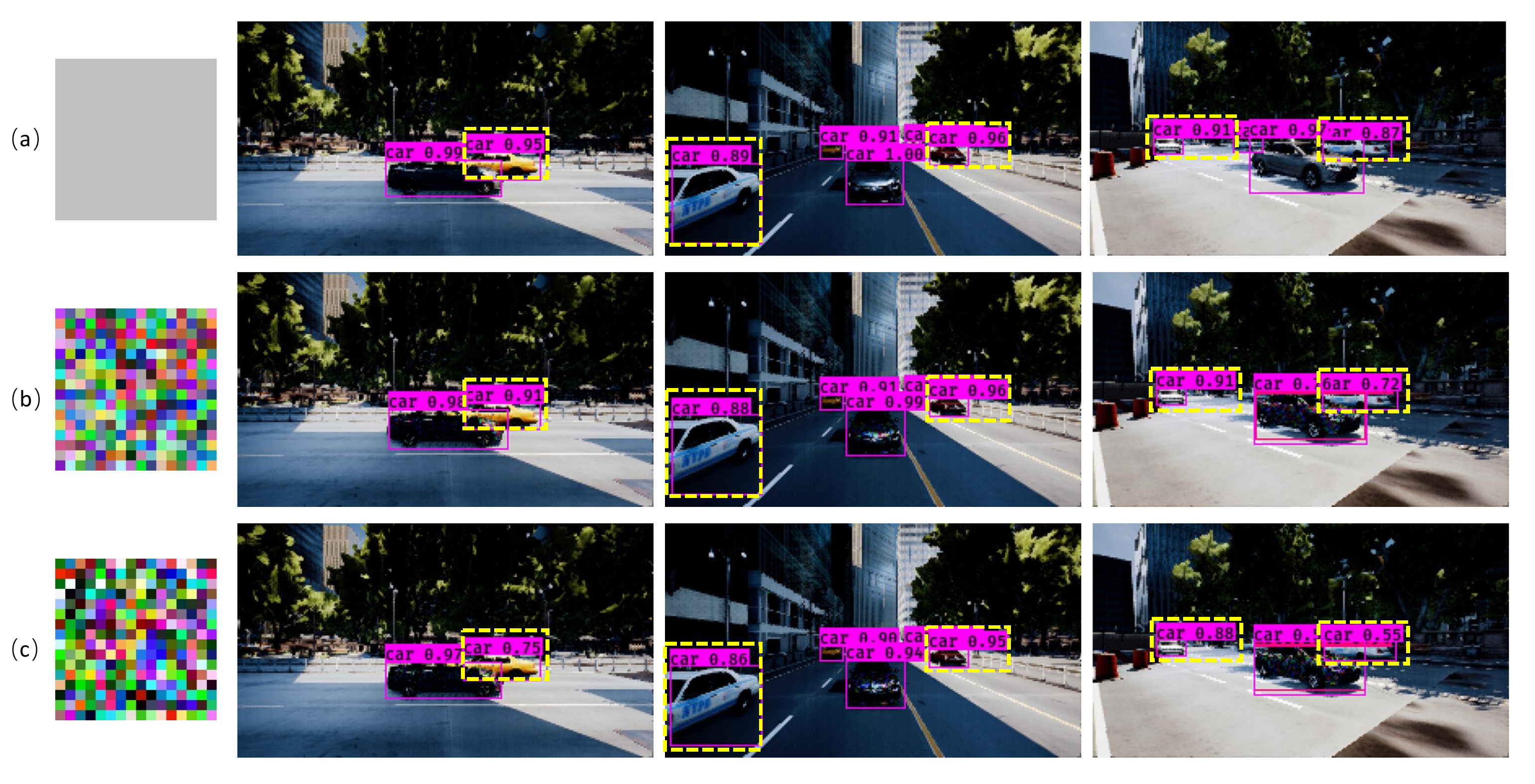}

\caption{YOLOv3 detection visualization of a Toyota Camry 2015 XLE with silver paint (a), random texture (b) and our learned camouflage (c) in the simulation. Our camouflage successfully attacks the detector's performance on the unpainted vehicles (framed by yellow boxes) compared with the random texture and plain color.}
\label{fig:intro}

\end{figure*}

Unlike previous adversarial attack works, in this paper we intend to attack object detectors' performance by learning a perturbation on the contextual object of the input image. Our approach hinges on an evolution search strategy \cite{wierstra2008natural} for the optimization of the non-differentiable objective function, which involves a complex 3D environmental simulation when mapping from an input (a camouflage pattern) to an output (the object detection accuracy). The simulated output images are shown in Fig.~\ref{fig:intro}. Given a batch of vehicles and a collection of simulated environments \cite{zhang2018camou}, our goal is to generate a camouflage for one of the vehicle and influence the detectors' performance on the other unpainted vehicles in those environments. To be noted, our algorithm is also different from the classic patch-based adversarial attacks \cite{liu2018dpatch, lee2019physical}. Instead of placing the patch in a random location within the image, we directly change the appearance of the object to investigate the effect on other object detection performance, which could be more powerful and general for different scenarios.


Many steps in this procedure are like black-boxes to us. Knowing the camouflage pattern, we are able to paint it onto the vehicle, but we have no idea how to write out a differentiable expression for the “painting function”. Similarly, we can drive the vehicle with the painted camouflage to various places in the environments and take pictures of the vehicle, but we cannot write out a differentiable “photographing function”. The list goes on. As a result, we have to explicitly deal with the black-box steps.

A clone function is employed in \cite{wierstra2008natural} to approximate the black-box steps. The authors propose to alternatively learn the clone function and search for the camouflage using this clone function. However, the quality of the camouflage highly depends on the performance of this clone function. When the two inter-dependent items are jointly learned, it is easy to be trapped at a poor local optimum. Besides, the simulators often introduce some randomness into the rendering procedure, making the clone function actually a mis-specified model. In this paper, we instead explicitly leverage the randomness of the simulator, along with the unknown black-box steps in the procedure from painting a camouflage onto a contextual vehicle to the results of detecting the other vehicles within the image. According to the evolution search strategy \cite{wierstra2008natural}, we are estimating the expected gradient of a distribution via dense sampling the distribution. The gradient error, which is caused by a noisy function, is naturally bounded during the optimization by the reciprocal of the square root of the size of the search distribution, thus eliminating the need for ad-hoc steps such as repeated queries. Since we directly estimate the expected gradient w.r.t the camouflage, there would not be local minimum deadlock as mentioned in \cite{zhang2018camou}.

To sum up, we make the following contributions:
\begin{itemize}
\item We demonstrate that, for the first time, by learning a camouflage on a contextual vehicle, we could attack the detectors' performance on the unpainted vehicles in the same image. Our learned camouflage is validated to work on all of the three state-of-the-art object detectors.

\item We proposed a new method that jointly models the transformation distribution and camouflage variations. 

\item In addition to the contextual adversarial attack, we demonstrate the applicability of our CCA model to enhance the performance of object detectors. 
\end{itemize}

 
\section{Related Work}

In this section, we introduce some related works which aim to defunctionalize CNNs. While other authors use adversarial learning for purposes different from ours, such as extracting private information \cite{tramer2016stealing}, we narrow our scope to those that aim to defunctionalize CNNs.

\subsection{Adversarial Attack and Blackbox Optimization}
Whether targeting a classification model or a detection model, adversarial machine learning (AML) against CNNs has seen great development since Szegedy et al.~\cite{szegedy2013intriguing} discovered that small perturbations could alter CNN predictions. A considerable amount of recent AML literature \cite{arnab2018robustness, athalye2018obfuscated, lu2017adversarial, xie2017adversarial} requires a target model to be transparent
to calculate the gradient, as knowing a numerical pathway between a perturbation and a model prediction grants an enormous optimization advantage. Such a pathway is not always available given that a target 
model is often a blackbox in real attacks. Various proposed blackbox attack methods rely solely on
query and target model output. Most of these are still based
on gradient descent. The difference lies in how they estimate a
perturbation gradient in the blackbox setting. ZOO \cite{chen2017zoo} estimates
it by coordinate descent. \cite{papernot2017practical} trained a substitute
model to mimic the target model behavior to obtain the gradient.
\cite{li2018nattack,ilyas2018black} estimate it via an evolutionary strategy. \cite{su2019one}
is based on differential evolution. \cite{zhang2018camou} shares the same task with us and they choose to minimize an end-to-end clone
network that is trained to directly estimate score from camouflage.

\subsection{Physical Adversarial Attack}
One of the sub-areas of AML is to generalize it to the
physical dimension. This is challenging because training in
silico often results in poor generalization to real environments.
The first attempt was by \cite{kurakin2016adversarial} and they produced perturbations that remain effective when printed on paper. \cite{eykholt2018robust} generalize a perturbation to fool a classifier of real stop sign images. However \cite{lu2017no, lu2017standard} found that \cite{eykholt2018robust}’s perturbation do not work on a real world detector. This is likely because detectors are more robust than classifiers since detectors must detect objects at multiple scales. Later, \cite{lu2017adversarial,chen2017attacking} proposed perturbations that could attack stop-sign detectors in the real-world. However, all the aforementioned works are perturbing detectors of stop signs, which are planar objects whose images, under changes in camera geometry, are related by linear 2D projective transformations. This is in contrast to nonplanar objects whose images, under changes in camera geometry, are related by more complex range-dependent nonlinear transformations. Perturbations are easily transformed via linear 2D projective transformation without breaking the gradient chain between the perturbation and the output score \cite{46561,eykholt2018robust}. Complex nonlinear transformations, however, will require a dedicated 3D simulation, such as the one used in our paper, which breaks the gradient chain and greatly complicates the optimization. Besides, there are only 1983 stop sign instances in the training split of MS-COCO \cite{lin2014microsoft}. Meanwhile, there are 43867 cars of various appearance in it. Such a great amount and diversity benefits vehicle detectors’ generalization capacity trained on MS-COCO and makes it harder to fool.

\section{Approach}

\begin{figure*}[t]
\centering
\includegraphics[width=0.9\textwidth]{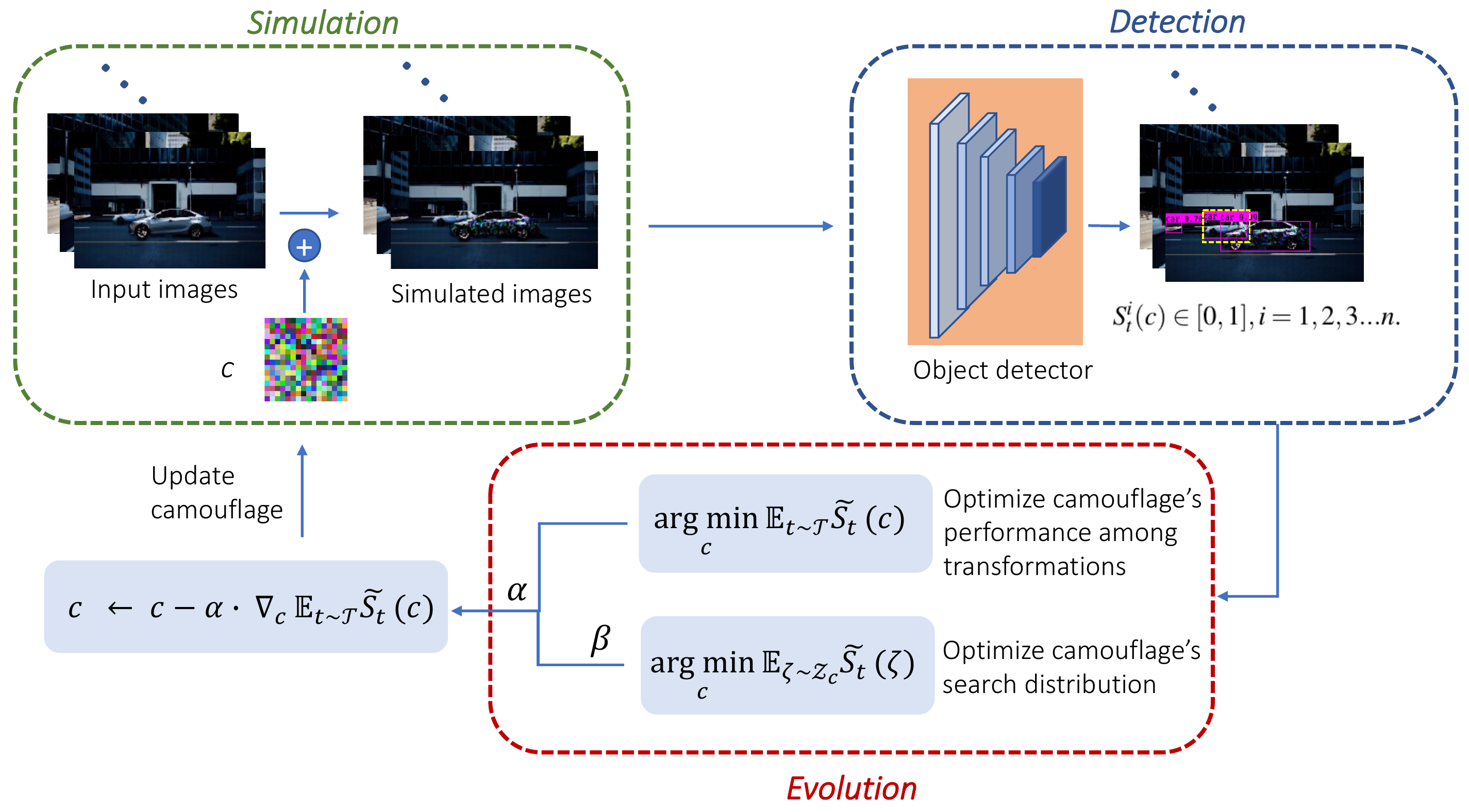}
\caption{The framework of proposed CCA algorithm.}
\label{fig:intro_0}
\end{figure*}

\subsection{Problem Statement}
\label{section:problem}
The framework of proposed algorithm is shown in Fig.~\ref{fig:intro_0}. There are mainly three phases in our algorithm, including simulation, detection and evolution. Let $c$ denote a vehicle camouflage pattern represented by an RGB image. When $c$ has been painted onto the surface of a vehicle in a tiled style, state-of-the-art object detectors would completely or partially fail to detect/classify the vehicle regardless of observation angles, distances and environments. In this context, we aim to determine whether it is possible to create a camouflage pattern $c$, being painted on a given vehicle, could mislead detectors to make incorrect detection/classification on the other vehicles (without camouflage) in the same scene. After inputting the simulated images into detectors and outputting the detection results, we conduct the optimization of the proposed camouflage from two aspects, namely the attack performance of the camouflage on detectors and search distribution of the camouflage. The workflow of proposed contextual adversarial attack is shown in Algorithm 1. We describe our approach to implementing the proposed loss function in Section \ref{section:loss}. In Section \ref{section:evolutionary} we outline the architecture of our evolutionary model to generate the optimal camouflage pattern.

\begin{algorithm}[htbp]
\caption{Framework of the proposed method}
\label{alg:Framwork}
\begin{algorithmic}[1]
\Require
  Object detector;
  Vehicle camouflage painter;
  Learning rate, $\alpha$
\Ensure
  Optimal camouflage pattern, $c$;
\State 
  Randomly initialize vehicle camouflage pattern $c$;
\Repeat
\State Painting the $c$ onto a vehicle for each $t$ in $\tau$ with vehicle camouflage painter;
\State Calculating the mean score ${\mathbb{E}_{t\sim\mathcal{T}}\widetilde{S}{_{t}}(c)}$ with object detector;
\State Calculating the gradient $\nabla_{c}\widetilde{S}_{t}(c)$;
\State Updating the search distribution $\mathcal{Z}_{c}$;
\State Updating $c$ according to $\nabla_{c}\mathbb{E}_{t\sim\mathcal{T}}\widetilde{S}_{t}(c)$ with learning rate $\alpha$;
\Until{$\mathbb{E}_{t\sim\mathcal{T}}\widetilde{S}{_{t}}(c)$ no longer decreases}\\
\Return $c$
\end{algorithmic}
\end{algorithm}

\subsection{Loss Function Designation}
\label{section:loss}
Given a camouflage pattern $c$, represented by a $16 \times 16$ RGB image, we would like to convert it to a photo that looks like real. We denote $t$ as the transformation, specifically including tiling and painting the $c$ onto a vehicle, moving the vehicle to a location in an environment and at last taking a photo of that vehicle from a certain angle and distance. After transformation, we will have an image with a vehicle covered by a camouflage texture in it. We suppose that there are $n$ vehicles in the generated image except for the one with camouflage. Then, we could denote the scores of the vehicles without camouflage as $S{^i_{t}}(c)\in [0,1], i = 1,2,3...n$. The overeaching idea of our algorithm is to find an optimal camouflage, such that the mean score of $S{^i_{t}}(c)$ is low:

\begin{equation}
   \mathop{\arg\min}_{c}{\frac{1}{n}\sum_{i=1}^{n}S{^i_{t}}(c)}
\end{equation}

For the concern about the complexity of the physical world, following \cite{pmlr-v80-athalye18b}, we construct a batch of transformations $\mathcal{T}$ for each learning procedure, involving the factors of different lighting condition, and distance and viewpoint changes. Thus, by minimizing the score in expectation, we have the following optimization problem:

\begin{equation}
   \mathop{\arg\min}_{c}{\mathbb{E}_{t\sim\mathcal{T}}\widetilde{S}{_{t}}(c)}, \label{e2}
\end{equation}
where $\widetilde{S}{_{t}}(c)$ is the mean score of $S{^i_{t}}(c)$. And then we could have eq. \ref{e2} as:

\begin{equation}
   \mathop{\arg\min}_{c}{\frac{1}{|\mathcal{T}|}\sum_{t \in \mathcal{T}}\widetilde{S}{_{t}}(c)}
\end{equation}

\subsection{Evolutionary Model}
\label{section:evolutionary}
In this section, we aim to optimize a flattened $c$ during the training process. Specifically, it contains two major components. First one is the performance over various transformations. Besides, we also want to optimize $c$'s search distribution’s members’ performance. Suppose $ \mathcal{Z}_{c}$ is the search distribution of camouflage $c$, then, for a fixed transformation $t$, our goal is to optimize:
\begin{equation}
   \mathop{\min}_{c}\mathbb{E}_{\zeta\sim\mathcal{Z}_{c}}\widetilde{S}_{t}(\zeta)
\end{equation}

Inspired by \cite{berny2001statistical}, we design our gradient as:
\begin{equation}
\begin{aligned}
\nabla_{c}\widetilde{S}_{t}(c) 
&\approx \nabla_{c}\mathbb{E}_{\zeta\sim\mathcal{Z}_{c}}\widetilde{S}_{t}(\zeta) \\
&\approx \frac{1}{\lambda}\sum_{k=1}^{\lambda}\widetilde{S}{_{t}}(\zeta_{k}) \cdot \nabla_{c}log\pi(\zeta_{k}|c),
\end{aligned}
\end{equation}
where $\lambda$ is the number of samples from the distribution. However, different from the problem mentioned in \cite{wierstra2008natural}, our optimization has a constraint of $[0, 255]$ because of the image feature of $c$. In our case, we propose to use a truncated multivariate normal distribution $\mathcal{N}_{[0:255]}$ to bound both $c$ and the search radius without extra hassle. 

Instead of directly minimizing $\mathbb{E}\widetilde{S}_{t}(c)$, we choose to minimize the binary cross-entropy between it and zero as shown eq. \ref{e4}:
\begin{equation}
H[0, \widetilde{S}_{t}(c)] = -log(1-\widetilde{S}_{t}(c)) \label{e4}
\end{equation} 

During optimization, we assume the scores of $\mathcal{Z}_{c}$ are normally distributed given that $\mathcal{Z}_{c}$ itself is normally distributed. In this case, by normalizing ${\mathbb{E}_{t\sim\mathcal{T}}\widetilde{S}{_{t}}(c)}$ into standard scores using $\beta_{t}$, we could have:
\begin{equation}
\begin{aligned}
&\beta_{k}=\frac{\mathbb{E}_{t\sim\mathcal{T}}\widetilde{S}{_{t}}(\zeta_k)- \frac{1}{\lambda}\sum_{k=1}^{\lambda}\mathbb{E}_{t\sim\mathcal{T}}\widetilde{S}{_{t}}(\zeta_k)}{std\left\{\mathbb{E}_{t\sim\mathcal{T}}\widetilde{S}{_{t}}(\zeta_n)|1\leq n \leq \lambda \right\}} \\
&\mathbb{E}_{t\sim\mathcal{T}}\widetilde{S}{_{t}}(c) = {\frac{1}{|\mathcal{T}|}\sum_{t \in \mathcal{T}}\widetilde{S}{_{t}}(c)}
\end{aligned}
\end{equation} 

Then, we arrive at our normalized gradient:
\begin{equation}
\begin{aligned}
&\nabla_{c}\mathbb{E}_{t\sim\mathcal{T}}\widetilde{S}_{t}(c) \\
&\approx \nabla_{c}\mathbb{E}_{(t, \zeta)\sim(\mathcal{T}, \mathcal{N}_{[0:255]}(c,\sigma^{2}) )}\widetilde{S}_{t}(\zeta) \\
&\approx \frac{1}{\lambda \cdot |\mathcal{T}|} \sum_{t \in \mathcal{T}} \sum_{k=1}^{\lambda} \beta_{k} \cdot H[0, \widetilde{S}_{t}(\zeta_{k})] \cdot \nabla_{c}log \pi(\zeta_{k}|c)\\
& = \frac{1}{\lambda \cdot |\mathcal{T}| \cdot \sigma^{2}}\sum_{t \in \mathcal{T}} \sum_{k=1}^{\lambda} \beta_{k} \cdot H[0, \widetilde{S}_{t}(\zeta_{k})] \cdot(\zeta_{k}-c), \label{e8}
\end{aligned}
\end{equation}

Then, the gradient approximation in eq.~\ref{e8} is used to perform gradient descent. We choose to perform simple gradient
descent which is:
\begin{equation}
c \leftarrow c - \alpha \cdot \nabla_{c}\mathbb{E}_{t\sim\mathcal{T}}\widetilde{S}_{t}(c)
\end{equation}
where $\alpha$ is the learning rate. Then we update $c$ until $\mathbb{E}_{t\sim\mathcal{T}}\widetilde{S}{_{t}}(c)$ no longer decreases. 







\begin{figure*}[htbp]
\centering
\includegraphics[width=0.9\textwidth]{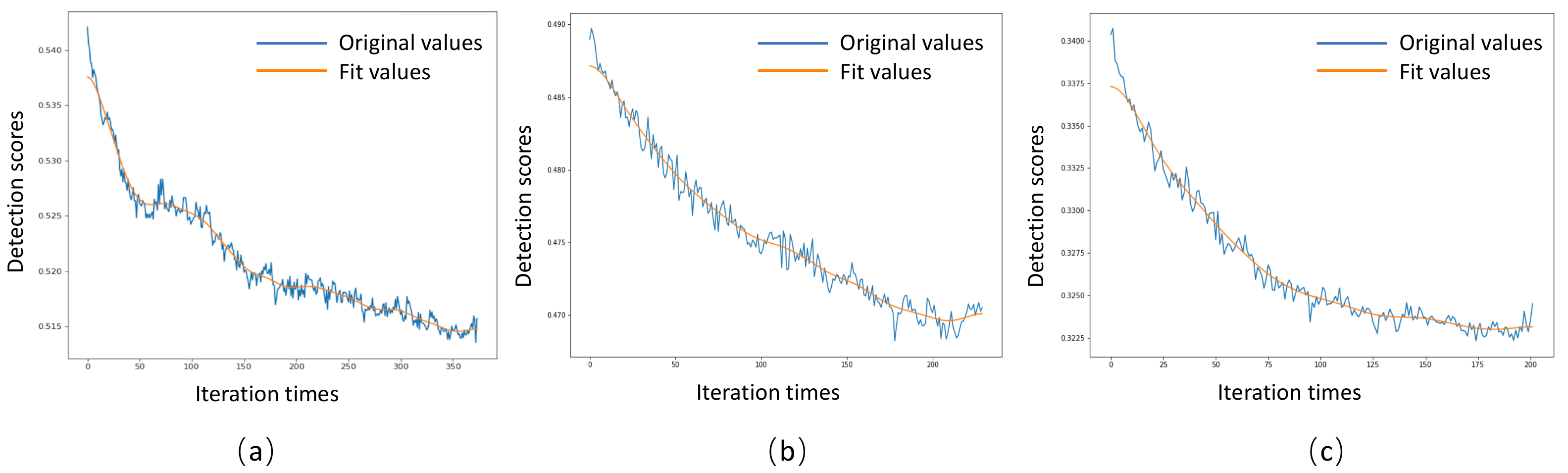}
\caption{Training process of our learned camouflage against state-of-the-art object detectors. The x axis represents the iteration time of the training, and the y axis represents the mean value of the detection scores of the unpainted vehicles in all scenes. (a) Training process against YOLOv3; (b) Training process against MaskRCNN; (c) Training process against FCOS.}
\label{fig:train}
\end{figure*}

\begin{table*}[htbp]
\centering
\begin{tabular}{l|lll|lll}
\hline
\multicolumn{1}{c|}{\multirow{2}{*}{Camouflages}} & \multicolumn{3}{c|}{Training set} & \multicolumn{3}{c}{Testing set} \\ \cline{2-7} 
\multicolumn{1}{c|}{} &
  \multicolumn{1}{c}{\begin{tabular}[c]{@{}c@{}}Detection \\ confidence(\%)\end{tabular}} &
  \multicolumn{1}{c}{mIOU(\%)} &
  \multicolumn{1}{c|}{P@0.5(\%)} &
  \multicolumn{1}{c}{\begin{tabular}[c]{@{}c@{}}Detection \\ confidence(\%)\end{tabular}} &
  \multicolumn{1}{c}{mIOU(\%)} &
  \multicolumn{1}{c}{P@0.5(\%)} \\ \hline
Basic colors                                    & 55.87       & 49.93      & 65.71      & 57.10      & 51.92      & 63.23     \\
Random camouflage                                      & 55.02       & 48.78      & 63.70      & 54.06      & 48.57      & 61.77     \\
Ours                                              & \textbf{51.35}       & \textbf{47.25}      & \textbf{62.35}      & \textbf{53.25}      & \textbf{48.15}      & \textbf{60.48}     \\ \hline
\end{tabular}
\caption{YOLOv3 detection performance among baselines and proposed CCA camouflage.}
\label{tab:table1}
\end{table*}

\begin{table*}[htbp]
\centering
\begin{tabular}{l|lll|lll}
\hline
\multicolumn{1}{c|}{\multirow{2}{*}{Camouflages}} & \multicolumn{3}{c|}{Training set} & \multicolumn{3}{c}{Testing set} \\ \cline{2-7} 
\multicolumn{1}{c|}{} &
  \multicolumn{1}{c}{\begin{tabular}[c]{@{}c@{}}Detection \\ confidence(\%)\end{tabular}} &
  \multicolumn{1}{c}{mIOU(\%)} &
  \multicolumn{1}{c|}{P@0.5(\%)} &
  \multicolumn{1}{c}{\begin{tabular}[c]{@{}c@{}}Detection \\ confidence(\%)\end{tabular}} &
  \multicolumn{1}{c}{mIOU(\%)} &
  \multicolumn{1}{c}{P@0.5(\%)} \\ \hline
Basic colors                                    & 48.48       & 46.61      & 49.78      & 48.87      & 47.54      & 48.19     \\
Random camouflage                                      & 48.90       & 47.08      & 49.61      & 48.79      & 47.46      & 47.87     \\
Ours                                              & \textbf{46.82}       & \textbf{45.54}      & \textbf{47.77}      & \textbf{47.97}      & \textbf{46.91}      & \textbf{46.71}     \\ \hline
\end{tabular}
\caption{MaskRCNN detection performance among baselines and proposed CCA camouflage.}
\label{tab:table2}
\end{table*}

\begin{table*}[htbp]
\centering
\begin{tabular}{l|lll|lll}
\hline
\multicolumn{1}{c|}{\multirow{2}{*}{Camouflages}} & \multicolumn{3}{c|}{Training set} & \multicolumn{3}{c}{Testing set} \\ \cline{2-7} 
\multicolumn{1}{c|}{} &
  \multicolumn{1}{c}{\begin{tabular}[c]{@{}c@{}}Detection \\ confidence(\%)\end{tabular}} &
  \multicolumn{1}{c}{mIOU(\%)} &
  \multicolumn{1}{c|}{P@0.5(\%)} &
  \multicolumn{1}{c}{\begin{tabular}[c]{@{}c@{}}Detection \\ confidence(\%)\end{tabular}} &
  \multicolumn{1}{c}{mIOU(\%)} &
  \multicolumn{1}{c}{P@0.5(\%)} \\ \hline
Basic colors                                    & 34.44       & 38.96      & 46.46      & 34.02      & 39.48      & 46.31    \\
Random camouflage                                      & 34.08       & 38.75      & 45.78      & 33.77      & 39.36     & 46.39     \\
Ours                                              & \textbf{33.02}       & \textbf{37.81}      & \textbf{44.83}      & \textbf{33.60}      &\textbf{38.53}      & \textbf{45.42}     \\ \hline
\end{tabular}
\caption{FCOS detection performance among baselines and proposed CCA camouflage.}
\label{tab:table4}
\end{table*}

\section{Experiments}

In this section, we present the experiment results to demonstrate the effectiveness of the proposed method. 

\subsection{Dataset} 
\label{data}
In this work, we use the Unreal Engine to build a simulation environment based on the real downtown environment. Specifically, we build different kind of cars, roads, traffic signals in the simulated environment. To obtain valid data for training and testing, we sequentially camouflage and place a professionally modeled Camry 2015 XLE in different locations in the environments. Also, 36 locations are sampled along the roads in the environment, 18 of which for training and the remaining 18 for testing. At each location, we record RGB images and groundtruth of the vehicle and the surrounding area from 20 fixed camera orientations. We manually label all the groundtruth of vehicles into two categories: the one carrying the camouflage and the normal ones. These images and groundtruths are then sent to the detector for evaluation. Inspired by \cite{zhang2018camou}, in our experiments, we set the resolution of camouflage to be $16\times16$.

\begin{figure*}[htbp]
\centering
\includegraphics[width=0.8\textwidth]{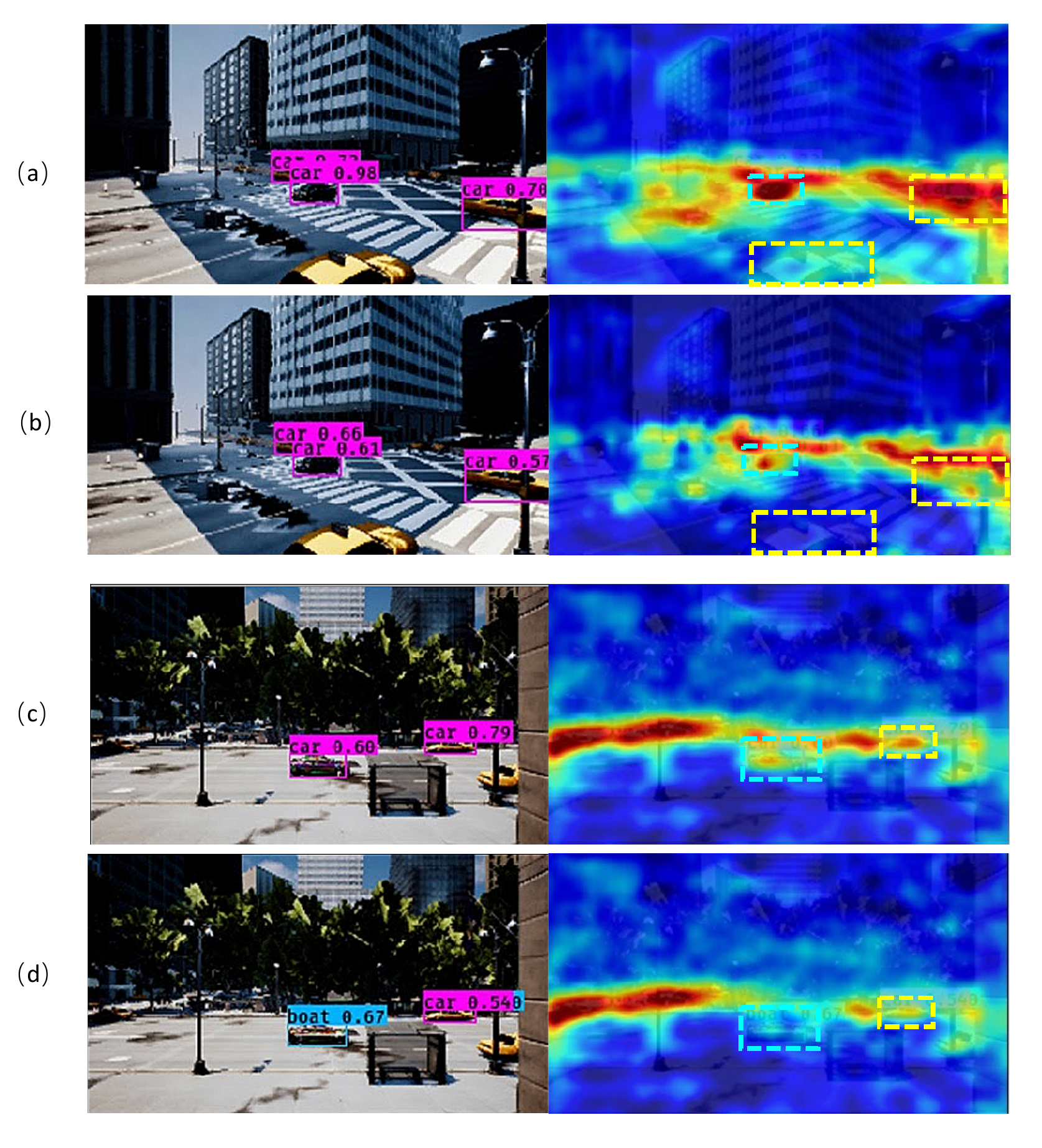}
\caption{Grad-CAM of YOLOv3 on images with random camouflage ((a)$\setminus$(c)) and CCA camouflage ((b)$\setminus$(d)). Left column is original images, and right column is Grad-CAM of images. In Grad-CAMs, pixels' color closer to red means more relevant to the detected category "car", and vice versa. The attentions on unpainted vehicles (framed by yellow boxes) are reduced with the proposed CCA camouflage being painted on the central vehicles.}
\label{fig:cam1}
\end{figure*}

\subsection{Implementation Details}
For evaluation, we use two MS-COCO \cite{lin2014microsoft} pretrained detectors in our experiments: YOLOv3 \cite{redmon2018yolov3} and Mask-RCNN \cite{he2017mask}. Mask-RCNN is currently one of the hallmark detectors for object detection in terms of detection performance. YOLOv3, on the other hand, balances performance with detection speed. We use the YOLOv3-SPP variant, which has the best detection performance among the YOLOv3 family. 

Also, we apply three detection metrics as the indicators of performance: detection confidence, mean Intersection over Union (mIoU) \cite{everingham2015pascal} and precision@0.5 (P@0.5). The detection confidence is the mean value of the scores of the detected cars in different scenes. The Intersection over Union $IoU$ between predicted bounding box $B_{p}$ and ground truth bounding box $B_{gt}$ is defined by $IoU =  \frac{B_{p}\cap B_{gt}}{B_{p}\cup B_{gt}}$  \cite{everingham2015pascal}. In our experiments, we use the best IoU achieved among all vehicle bounding box predictions for the image as its IoU. The mIoU is the averaged IoU across all images at all locations that are being reported. The P@0.5 is the percentage of images that have IoU larger than 0.5. This metric is used in the PASCAL detection challenge \cite{everingham2015pascal}, which considers a detection prediction to be successful if its IoU is greater than 0.5.

For a thorough evaluation, inspired by \cite{zhang2018camou}, we compare our results with two baselines to validate the effective of the proposed camouflage, as shown in Fig.~\ref{fig:intro}. The first one is conducted by painting the central vehicle with six real vehicle colors (red, black, silver, grey, blue, and white) and compute the mean value of the detection metrics. Another one is that we generate 5 random camouflages with the same resolution of the learned camouflage and report the mean values of the detection metrics for comparison.

\begin{figure}[htbp]
\centering
\includegraphics[width=0.48\textwidth]{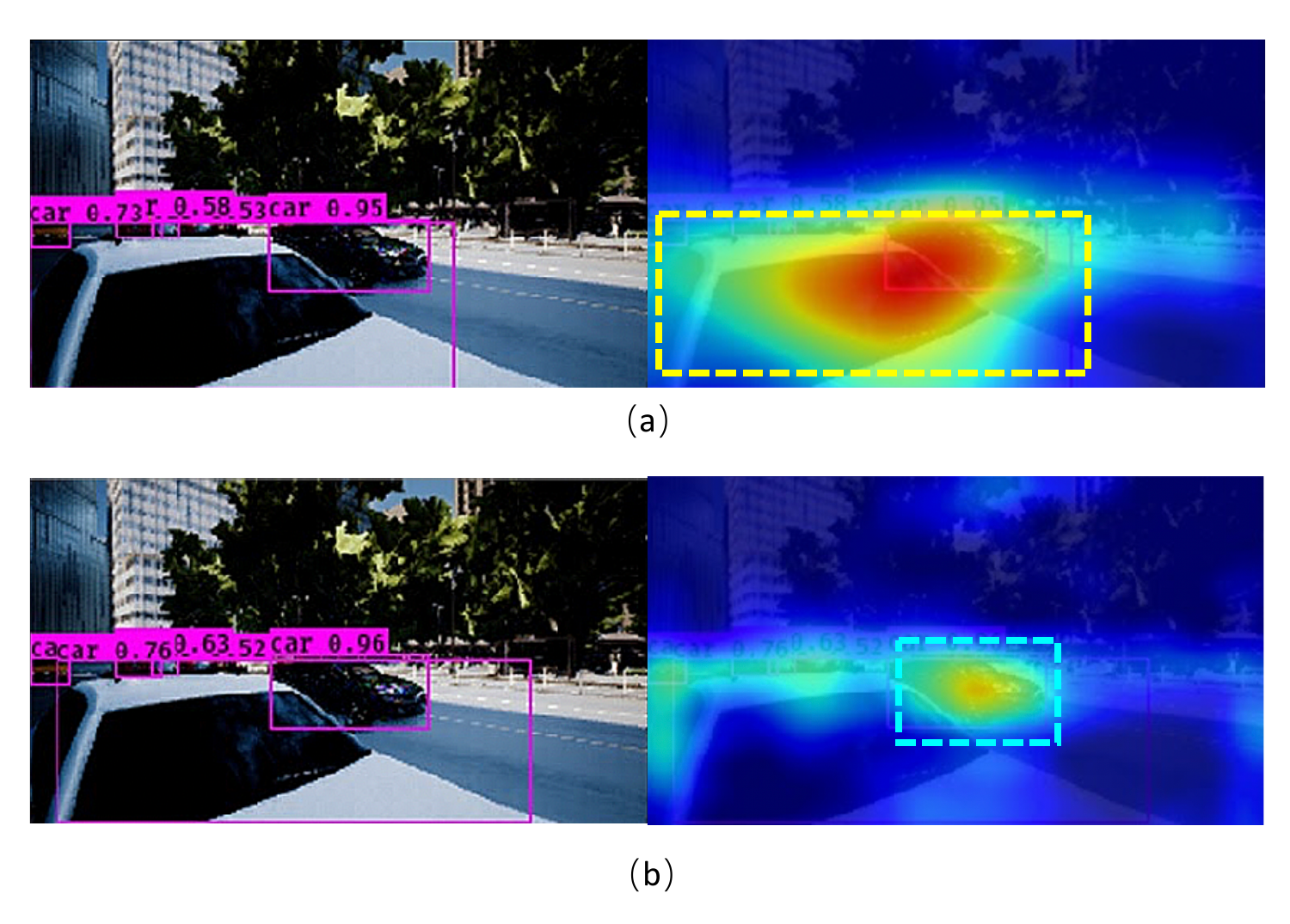}
\caption{Grad-CAM of YOLOv3 on images with random camouflage and CCA camouflage. (a) is image with random camouflage; (b)is image with learned CCA camouflages. Left row shows original images, and right row shows Grad-CAM of images. Attentions on the unpainted vehicle (framed by yellow box) are transferred to the camouflaged vehicle(framed by blue box) after training.}
\label{fig:cam2}
\end{figure}

\begin{figure*}[htpb]
\centering
\includegraphics[width=0.9\textwidth]{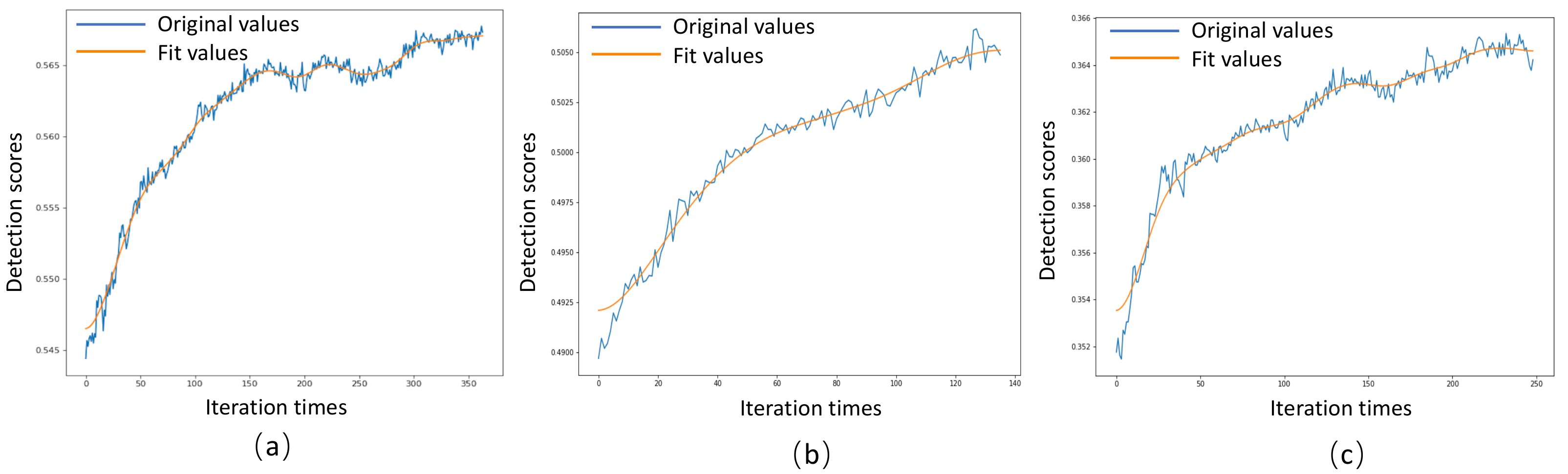}
\caption{Training process of our learned camouflage enhancing state-of-art object detectors. The x axis represents the iteration of the training, and the y axis represents the mean value of the detection scores of the unpainted vehicles in all scenes. (a) Training process of YOLOv3; (b) Training process of MaskRCNN; (c) Training process of FCOS.}
\label{fig:train2}
\end{figure*}

\subsection{Against Object Detectors} 
In this section, we present the performance of our learned camouflage and the baselines against several state-of-the-art detectors, including YOLOv3, Mask-RCNN and FCOS in the downtown environment.

The learning process of our CCA camouflage against YOLOv3 is shown in Fig. ~\ref{fig:train} (a). In the training, we set the scores by computing the mean value of the detection scores of all the unpainted vehicles. From Fig. ~\ref{fig:train} (a), 
we can easily observe that the detection scores of the vehicles decreases under the proposed learning algorithm, which demonstrates the effectiveness of the evolution search strategy we applied in CCA.
To further validate the effectiveness of the learned camouflage, experiments were conducted to compare the proposed camouflage to other baselines: painting basic colors or random camouflage on the context vehicle. As shown in Table~\ref{tab:table1}, our method outperforms baselines over all the three metrics on both training data and testing data. This supports our argument that we could misled object detector by only changing a part of the context. 

We present the training process and camouflages’ performance against MaskRCNN in Fig.~\ref{fig:train}(b) and Table~\ref{tab:table2}. Similar to its performance against YOLOv3, our learnd CCA camouflage outperforms baseline colors and random camouflages on both training and testing data. Furthermore, we can draw the same observation from the experiment results of FCOS reported in Fig.~\ref{fig:train}(c) and Table~\ref{tab:table4}. All the reported experiments show the effectiveness of CCA on attacking state-of-the-art object detectors.

\subsection{Qualitative Analysis}
To validate the proposed CCA camouflage's influence on object detectors, we visualize the Grad-CAM \cite{selvaraju2017grad} maps of the samples with our learned camouflage and random camouflage. Since the Grad-CAM shows the gradient heatmap of the classification score scalar w.r.t. the input, it could help us to identify the image regions most relevant to the particular category. Fig.~\ref{fig:cam1} shows two visualization examples, and each of them presents the detection result of YOLOv3 and corresponding Grad-CAM map. To be noted, YOLOv3 is a multi-scale detector and there are three layers of features with different sizes in the output. Considering of the resolution of the input and the objects' size in the images, we use the nearest convolution layer to the $13 \times 13$ feature layer of YOLOv3 as the penultimate layer and output the attention map of strategy 'car' in each detection. For a fair comparison, we output the absolute value instead of the normalized value of the gradient scores of the feature map in our experiments. As expected, in the images with the proposed CCA camouflaged vehicle (Fig.~\ref{fig:cam1} (b)(d)), the unpainted vehicles gain much lower attention than in the scenes with a random camouflaged vehicle (Fig.~\ref{fig:cam1} (a)(c)). This observation further explains and validates our assumption that we could reduce the object detectors' performance on certain objects by learning a contextual camouflage. 

Fig. ~\ref{fig:cam2} shows another example, in which the detector's attention is completely shifted from the front unpainted vehicle to the vehicle in the back. This case also proves the the proposed CCA camouflage's capability of influencing the detector's performance on the objects that without camouflages painted on them.

\subsection{Extensions on Contextual Adversarial Attack}
So far we have shown the effectiveness of our learned CCA camouflage on attacking object detector with the evolution strategy. Intuitively, we come up with the assumption that it should also be possible that we influence the detector with an opposite effected camouflage and increase the attention of the vehicles.

Here, we confirm that by experiments. In order to gain the opposite rewards, we reset the loss function proposed in Section ~\ref{section:loss} as:
\begin{equation}
   \mathop{\arg\min}_{c}{-\frac{1}{|\mathcal{T}|}\sum_{t \in \mathcal{T}}\widetilde{S}{_{t}}(c)}.
\end{equation}

Then we apply the evolution strategy proposed in Section ~\ref{section:evolutionary} to figure out the potential positive affect of the CCA camouflage on object detectors. Fig. ~\ref{fig:train2} shows the training process of the YOLOv3 and MaskRCNN. It shows that we can effectively improve the performance of the models with a learned contextual camouflage. Detection results comparison also proves this as shown in Table ~\ref{tab:table3}. The learned CCA camouflage outperforms the basic colors and random camouflage both on training and testing dataset. That means we could influence the object detectors' performance by not only attacking them but also enhancing them with the proposed CCA pattern. 

\begin{table*}[]
\setlength{\abovecaptionskip}{20pt}
\centering
\begin{tabular}{l|l|lll|lll}
\hline
\multirow{2}{*}{Networks}   & \multicolumn{1}{c|}{\multirow{2}{*}{Camouflages}} & \multicolumn{3}{c|}{Training set} & \multicolumn{3}{c}{Testing set} \\ \cline{3-8} 
 &
  \multicolumn{1}{c|}{} &
  \multicolumn{1}{c}{\begin{tabular}[c]{@{}c@{}}Detection \\ confidence(\%)\end{tabular}} &
  \multicolumn{1}{c}{mIOU(\%)} &
  \multicolumn{1}{c|}{P@0.5(\%)} &
  \multicolumn{1}{c}{\begin{tabular}[c]{@{}c@{}}Detection \\ confidence(\%)\end{tabular}} &
  \multicolumn{1}{c}{mIOU(\%)} &
  \multicolumn{1}{c}{P@0.5(\%)} \\ \hline
\multirow{2}{*}{YOLOv3} & Random camouflage                                 & 55.02       & 48.78      & 63.70      & 54.06      & 48.57      & 61.77     \\
                            & CCA camouflage                                & \textbf{56.77}       & \textbf{50.09}      & \textbf{64.94}      & \textbf{54.97}      & \textbf{49.90}      & \textbf{61.89}     \\ \hline
\multirow{2}{*}{MaskRCNN}   & Random camouflage                                 & 48.90       & 47.08      & 49.61      & 48.79      & 47.46      & 47.87     \\
                            & CCA camouflage                                & \textbf{50.61}       & \textbf{47.43}      & \textbf{51.09}      & \textbf{49.30}      & \textbf{47.86}      & \textbf{49.04}     \\ \hline
\multirow{2}{*}{FCOS}   & Random camouflage                                 & 34.08       & 38.75      & 45.78      &33.77      & 39.36     & 46.39     \\
                            & CCA camouflage                                & \textbf{35.21}       & \textbf{39.70}      & \textbf{46.74}      & \textbf{34.46}      & \textbf{39.97}      & \textbf{46.47}     \\ \hline
\end{tabular}
\caption{Detection performance of state-of-the-art object detectors on camouflages learned with opposite rewards.}
\label{tab:table3}
\end{table*}

\section{Conclusion}
In this paper, we first investigate the problem of learning contextual adversarial object camouflage to attack vehicle detectors, such as YOLOv3, MaskRCNN and FCOS. We propose an evolutionary based algorithm to learn highly effective camouflages by interacting with a photo-realistic simulation. Our proposed CCA algorithm not only shows the effectiveness of attacking the state-of-the-art object detectors, but also shows its capability to enhance the detectors. The next phase of our work is to generalize the camouflages from simulation to the real world. Both adversarial domain adaptation and domain randomization seem to be promising approaches for this step.

\bibliographystyle{IEEEtran}
\bibliography{bare_conf}
%




\end{document}